\documentclass[wcp]{jmlr}

\usepackage{amsmath}
\usepackage{algorithm}
\usepackage{fancyvrb}
\usepackage{multicol}
\usepackage{comment}
\usepackage{mdframed}
\usepackage{enumitem}
\usepackage{relsize}
\usepackage{multicol}
\usepackage{framed}

\setlist{nolistsep}
\usepackage{algpseudocode}
\usepackage{graphicx}

\newcommand{\welllogscale}{0.45 } %0.55

\jmlrvolume{1}
\jmlryear{2014}
\jmlrworkshop{ICML 2014 AutoML Workshop}

\title[inverting Sequential Simulators Using Probabilistic Programming]{Automatic Inference for Inverting Software Simulators via Probabilistic Programming}

  \author{\Name{Ardavan Saeedi} \Email{ardavans@mit.edu}\\
   \Name{Vlad Firoiu} \Email{vladfi@mit.edu}\\
   \Name{Vikash Mansinghka} \Email{vkm@mit.edu}}

\begin{document}

\maketitle
\vspace{-0.5in}
\begin{abstract}
Models of complex systems are often formalized as sequential software simulators: computationally intensive programs that iteratively build up probable system configurations given parameters and initial conditions. These simulators enable modelers to capture effects that are difficult to characterize analytically or summarize statistically. However, in many real-world applications, these simulations need to be inverted to match the observed data. This typically requires the custom design, derivation and implementation of sophisticated inversion algorithms. Here we give a framework for inverting a broad class of complex software simulators via probabilistic programming and automatic inference, using under 20 lines of probabilistic code. Our approach is based on a formulation of inversion as approximate inference in a simple sequential probabilistic model. We implement four inference strategies, including Metropolis-Hastings, a sequentialized Metropolis-Hastings scheme, and a particle Markov chain Monte Carlo scheme, requiring 4 or fewer lines of probabilistic code each. We demonstrate our framework by applying it to invert a real geological software simulator from the oil and gas industry.
\end{abstract}

\section{Introduction}

Sequential software simulators are used to model complex systems in fields ranging from geophysics~\citep{symes2011modelling} to finance~\citep{calvet2007multifrequency}. They can capture dynamics that produce effects which are difficult or impossible to characterize analytically or to summarize statistically. However, the real-world problems faced by modelers often require inference, not just simulation. For example, prediction tasks require identifying realizations of a simulation that are compatible with observed data. The problem of identifying probable realizations of a simulator given data is sometimes called {\em simulator inversion}. Both deterministic, optimization-based methods~\citep{boschetti1996inversion},~\citep{ramillien2001genetic} and stochastic, sampling based ~\citep{malinverno2002parsimonious},~\citep{chen2006development} methods are sometimes applied. Applying a standard technique to a new simulator or developing a new method for an existing simulator requires developing and implementing custom algorithms.

In this paper we show how to use probabilistic programming and automatic inference to formulate and solve a broad class of inversion problems. We define a simple interface to a sequential software simulator, and define a probabilistic model and approximate inference problem for inversion given that interface. This formulation requires under 20 lines of probabilistic code. We also describe four Monte Carlo inference strategies for solving the inversion problem, each requiring 4 or fewer lines of probabilistic code. We demonstrate our framework by applying it to invert a real geological software simulator from the oil and gas industry.

\section{A framework for inverting sequential simulation software}
\label{section:framework}

To invert sequential simulators, we define a probabilistic model over their parameters that encodes generic priors, and use approximate inference methods to infer probable values given data.

\begin{comment}
The way we define the model has significant effect on the inference method that we use for the inversion.
\end{comment}

\begin{figure*}[!htb]
\vspace{-2mm}
\begin{tabular}{rr}
\includegraphics[trim = 20mm 41mm 48mm 27mm, clip, scale = 0.37]{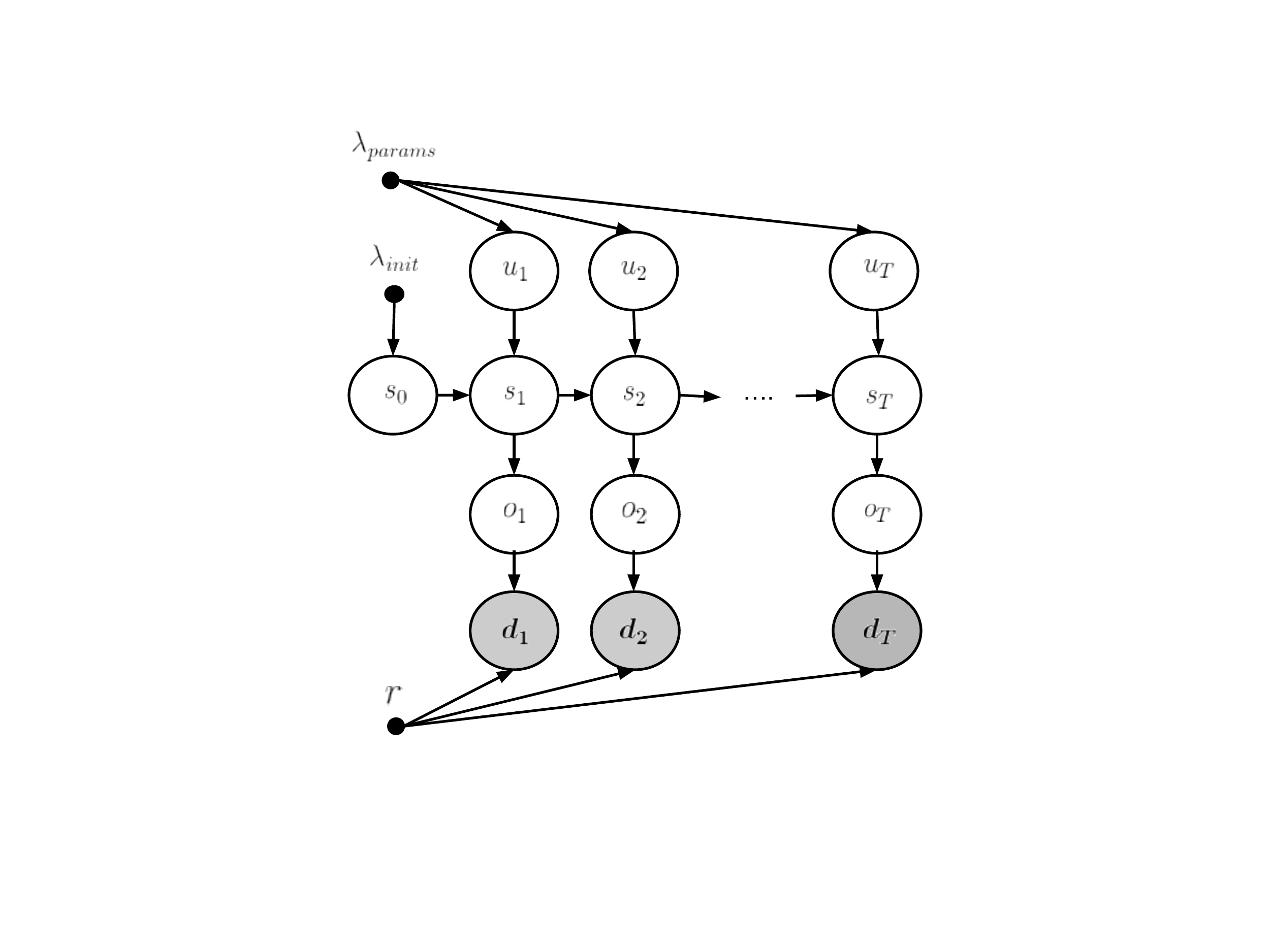} & 
\centering\fbox{

	\begin{minipage}[b][1.4in]{3in}
		\vspace{-3in}
\begin{enumerate}
\footnotesize{
\item \texttt{$s_0$ = Initialize($\lambda_{init}$)} and 

\texttt{$p_{init}(s_0;\lambda_{init})$ = Prob\_Init($s_0$,$\lambda_{init}$)}}
\item \texttt{$u_t$ = Sample($\lambda_{params}$)} and 

\texttt{$p_{params}(u_t;\lambda_{params})$ = Prob\_Smp($u_t$,$\lambda_{params}$)} 
  \item \texttt{$s_t$ = Simulate($s_{t-1}, u_t$)} and 

\texttt{$p_{sim}(s_{t};s_{t-1},u_{t})$ = Prob\_Sim($s_{t-1}$,$u_t$)}
  \item \texttt{$o_t$ = Emit($s_t$)} and 

\texttt{$p_{emit}(o_{t};s_{t})$ = Prob\_Emt($o_{t}$,$s_t$)}
\item \texttt{$k_\gamma(o_t, r)$ = Comp\_Distance\_Likelihood($o_t, r$)} 
\end{enumerate}
	\end{minipage}
	}
\end{tabular}
\vspace{-3mm}
\caption{\footnotesize{ \textbf{Our framework for inverting sequential simulators.} (\emph{left}) A probabilistic graphical model that describes the inference problem corresponding to simulator inversion. Each slice corresponds to a step in the sequential simulation, capturing the dependence of the new state on the previous state and new input parameters. See main text for more details. (\emph{right}) The procedural interface for specifying the simulator.}}
\label{figA:models}
\vspace{-7mm}
\end{figure*}

We assume the simulator is Markovian; that is, at every time point $t \in \{1, \dots, T\}$, we have a state variable $s_t$ which only depends on the previous state $s_{t-1}$ and the parameter(s) for the state $u_t$: $s_t|s_{t-1}, u_t \sim p_{sim}(s_{t-1}, u_t)$ and $u_t|\lambda_{params}\sim p_{params}(\lambda_{params})$. For the initial state, we assume $s_0|\lambda_{init}\sim p_{init}(\lambda_{init})$. Moreover, at every $t$ given the current state $s_t$, an emission $o_t$ is generated from a distribution, $o_t|s_t \sim p_{emit}(s_t)$. 
To afford the flexibility for many different forms of observable data, we allow simulators to come with arbitrary per-step likelihood terms. These terms are incorporated by defining a Bernoulli distribution with parameter $k_\gamma(o_t, r)$, where $r$ is the real data and $d_t|o_t, r, \gamma \sim Br(k_\gamma(o_t, r))$. We provide an example of the $k_\gamma(o_t, r)$ in Section~\ref{sec:likelihood}.

\begin{comment}
Finally, the joint probability distribution for the sequential model is:
\vspace{0.025mm}
\begin{align*}
p(\text{\bf{s, u , o, d}}|\lambda_{init}, \lambda_{params}, r, \gamma) = p_{init}(s_0|\lambda_{init})\prod^T_{t=1}p_{params}(u_t|\lambda_{params}) \times\\
p_{sim}(s_t|s_{t-1}, u_t) p_{emit}(o_t|s_t) Br(k_\gamma(o_t, r))
\end{align*}
\end{comment}

\begin{comment}
\begin{align*}
s_0|\lambda_{init}&\sim p_{init}(\lambda_{init})\\
u_t|\lambda_{params}&\sim p_{params}(\lambda_{params})\\
s_t|s_{t-1}, u_t &\sim p_{sim}(s_{t-1}, u_t)\\
o_t|s_t &\sim p_{emit}(s_t)\\ 
d_t|o_t, r, \gamma &\sim Br(k_\gamma(o_t, r))
\end{align*}
\end{comment}

\noindent

\subsection{Procedural interface for specifying sequential simulators}
\label{proc}

This probabilistic model lends itself to a natural software interface that can be satisfied by many sequential simulators: 
\vspace{3mm}
\begin{enumerate}
\item \texttt{$s_0$ = Initialize($\lambda_{init}$)} and \texttt{$p_{init}(s_0;\lambda_{init})$ = Prob\_Init($s_0$,$\lambda_{init}$)}: These procedures return the state of the simulator at initialization and compute the probability of sampling state $s_0$ from the initializing distribution respectively.
\vspace{0.02mm}
\item \texttt{$u_t$ = Sample($\lambda_{params}$)} and \texttt{$p_{params}(u_t;\lambda_{params})$ = Prob\_Smp($u_t$,$\lambda_{params}$)}: These procedures sample the parameters for a time point $u_t$ and compute the probability of sampling. 

  \item \texttt{$s_t$ = Simulate($s_{t-1}, u_t$)} and \texttt{$p_{sim}(s_{t};s_{t-1},u_{t})$ = Prob\_Sim($s_{t-1}$,$u_t$)}: Given the current state of the simulator at time $t$ and the parameter(s) for that time point, these procedures return the next state $s_t$ and compute the probability of sampling.
  
  \item \texttt{$o_t$ = Emit($s_t$)} and \texttt{$p_{emit}(o_{t};s_{t})$ = Prob\_Emit($o_{t}$,$s_t$)}: Given the current state at time $t$, these procedures emit the observation for that time point and computes the probability of emission.
\item \texttt{$k_\gamma(o_t, r)$ = Comp\_Distance\_Likelihood($o_t, r$)} : Given the observation and the real data, this procedure calculates the probability of having an observation at time $t$.
\end{enumerate}

\subsection{A real-world example: inverting a geological forward model}
\label{subsection:shell}

\begin{figure*}[!htb]
\vspace{-3.5mm}
\begin{tabular}{cc}

\includegraphics[trim = -15mm 73mm 15mm 40mm, clip, scale = 0.4]{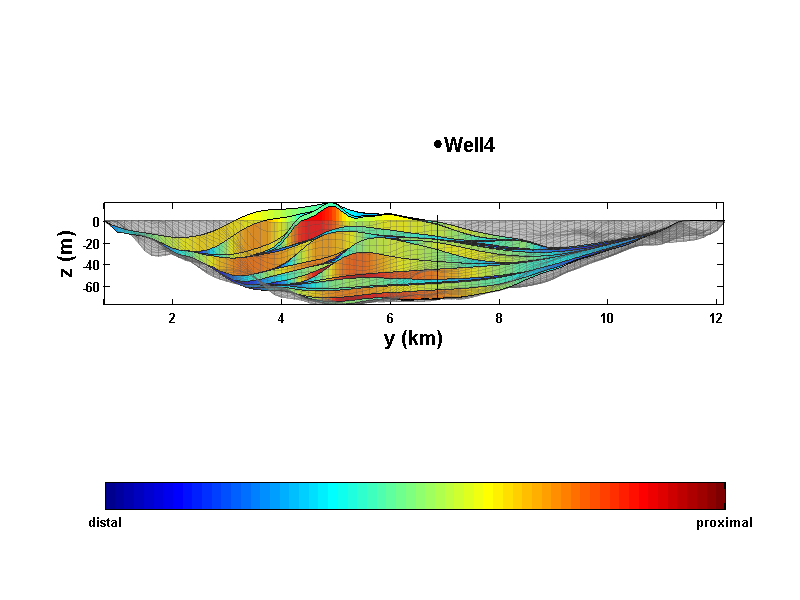} &
\includegraphics[trim = -20mm 10mm 100mm 90mm, clip, scale = 0.26]{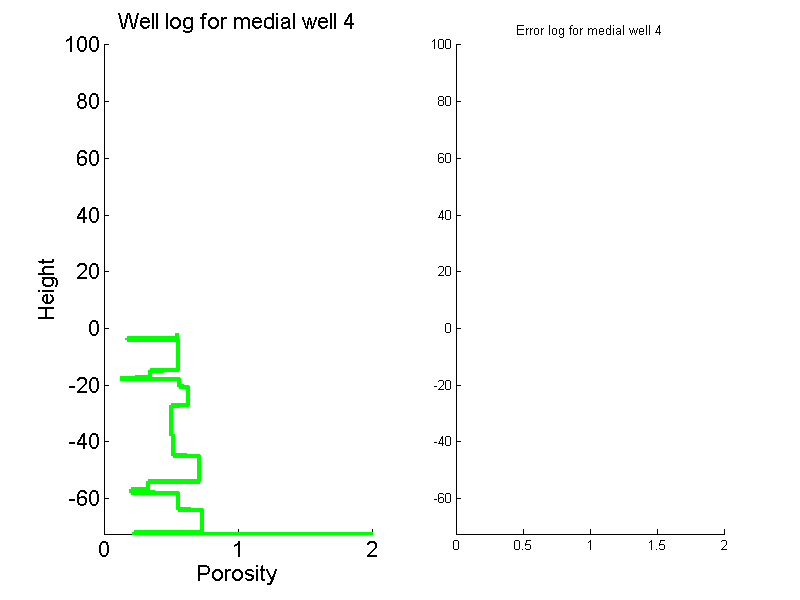}\\

\begin{comment}
\includegraphics[trim = -15mm 77mm 15mm 54mm, clip, scale = 0.5]{Figures/well4_iter20.png} &
\includegraphics[trim = -20mm 10mm 100mm 101mm, clip, scale = 0.35]{Figures/welllog4_iter20_n.png}\\
\end{comment}

\includegraphics[trim = -15mm 76mm 15mm 54mm, clip, scale = 0.4]{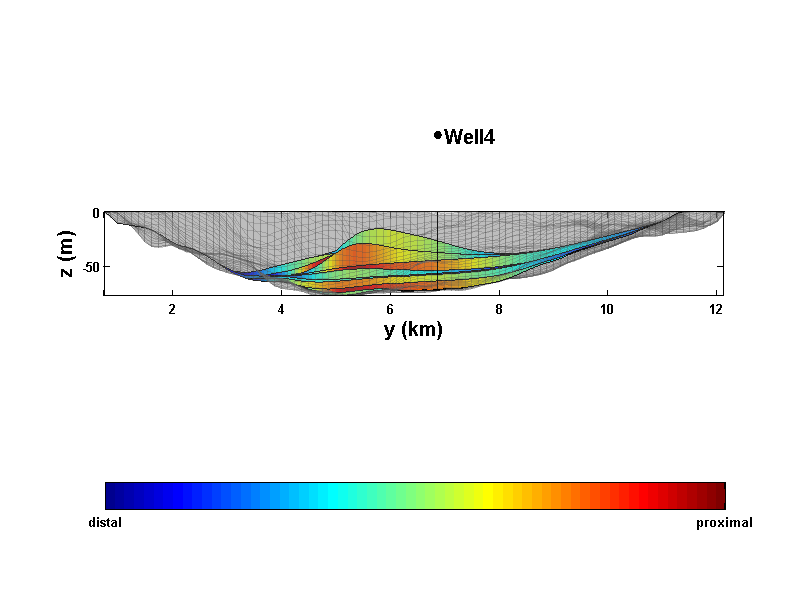} &
\includegraphics[trim = -20mm 10mm 100mm 101mm, clip, scale = 0.26]{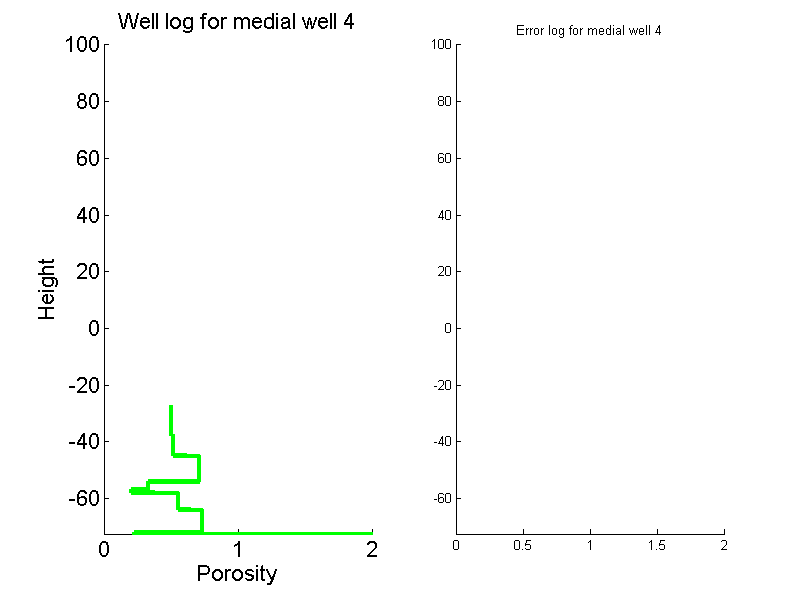}\\

\includegraphics[trim = -15mm 40mm 15mm 50mm, clip, scale = 0.4]{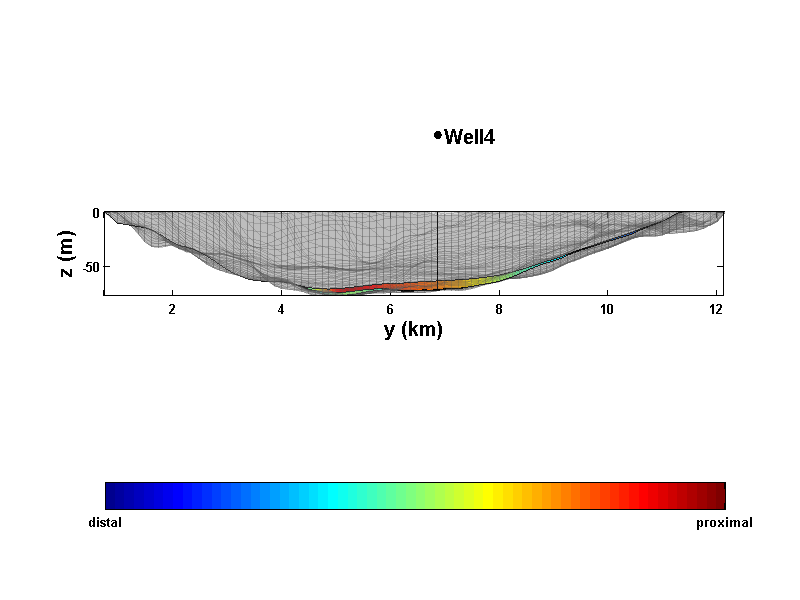} &
\includegraphics[trim = -20mm -37mm 100mm 101mm, clip, scale = 0.26]{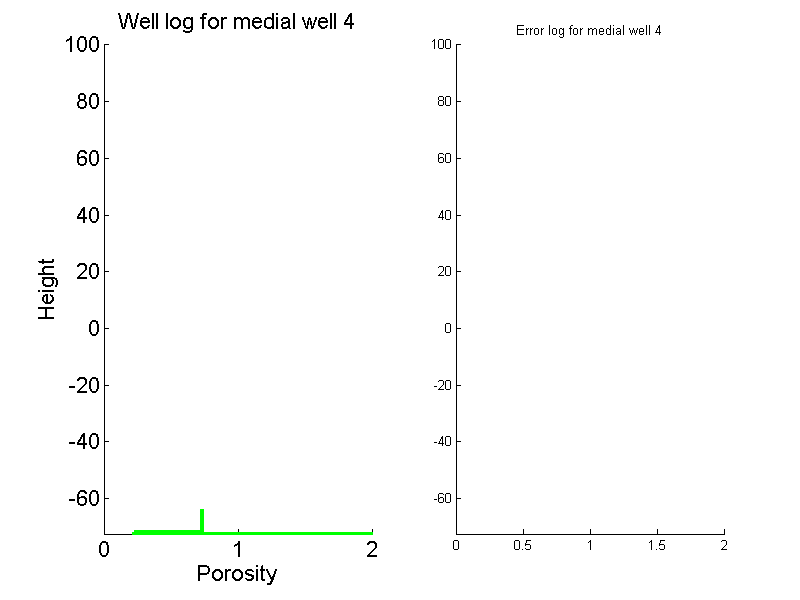}\\
\end{tabular}
\vspace{-15mm}
\caption{\footnotesize{\textbf{Lobe formation in the model.} (\emph{Left}) The side view of stratigraphy at a sample well location for sequential sampling of the lobes (for lobes 1, 10, and 30). (\emph{Right}) The generated well log for the same well. The x axis is the porosity value and the y axis is the height.}}
\label{fig:lobes}
\vspace{-8mm}
\end{figure*}

\begin{figure*}[!htb]
\begin{center}
\includegraphics[trim = 27mm 32mm 0mm 10mm, clip, scale = 0.48]{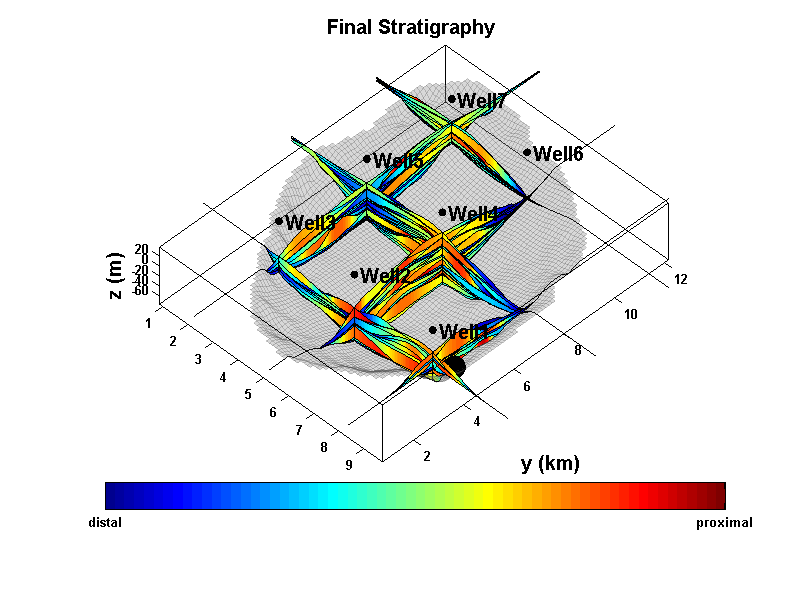}\\
\end{center}
\caption{\footnotesize{\textbf{Final output stratiagraphy,} showing the location of all 7 wells and many of the lobes.}}
\label{fig:strat}
\end{figure*}

We focus on a simulator developed by an oil and gas company. In this model, the states ($s_t$) correspond to geological features called a \emph{lobes} and the emissions ($o_{t}$) correspond to the porosities of the substrate. The parameters $u_t$ for each state are $n$-tuples of independent uniform random variables, $u_{t} \sim Unif[0,1]^{n}$. The real data $r$ is given by a set of $L$ \emph{well logs}, or sequences of porosities at varying heights, each at a different location $g_\ell$ in the geological model. Figure~\ref{fig:strat} shows a generated sample from the geological simulator. In our dataset, well logs are available for $L=7$ wells. The color of the lobe in renderings from the simulator represents the porosity at that lobe. See Figure~\ref{fig:lobes} for a visualization of lobe formation and Figure~\ref{fig:strat} for the final output stratigraphy showing all 7 wells.

The simulator builds a lobe $s_t$ according to a complex geological model $\Psi$ which given $s_{t-1}$ and $u_t$ is deterministic. The emission at a well location, denoted by $o_{t,\ell}$, is a function $\Phi$ of the current lobe $s_t$ and location $g_\ell$. We assume the initial state $s_0$ is a function of a hyperparameter $\lambda_{init}$ and the emissions at different locations are independent. For every emission $o_{t,\ell}$ \footnote{To be more precise, every emission $o_{t,\ell}$ at well $\ell$ is a sequence of porosities $\nu_{h, \ell}$ indexed by height $h$. The height at the end of lobe $t$ at well $\ell$ is denoted by $end_{t,\ell}$; hence, the generated observation at well $\ell$ and lobe $t$ is given by $o_{t,\ell}= (\nu_{end_{t-1,\ell},\ell},\nu_{end_{t-1,\ell}+1, \ell}, \dots, \nu_{end_{t,\ell}, \ell})$. Similarly, we can define the sequence of porosities for the real well logs.}, there will be a corresponding real well log for the same location and lobe $r_{t,\ell}$. We set $o_{t} = (o_{t,1},\dots,o_{t,L})$ and $r_t=(r_{t,1},\dots,r_{t,L})$.

The generative model for the observations at each lobe can be summarized as: 
\vspace{-1mm}
\begin{align*}
s_0|\lambda_{init} &\sim p_{init}(\lambda_{init}) \\
u_{t} &\sim Unif[0,1]^{n}\\
s_{t}|s_{t-1}, u_{t} &\sim \delta_{\Psi(s_{t-1}, u_{t})}\\
o_{t,\ell}|s_{t} &\sim \delta_{\Phi(s_{t}, g_\ell)}\\
d_t|o_t, r, \gamma &\sim Br(k_\gamma(o_t, r))
\end{align*}

The problem of stochastic inversion of a sequential simulator is finding the joint posterior of the states and the parameters given a sequence of real data $r$ (i.e. finding $p(u_{1:t}, s_{1:t}, o_{1,t}|d_{1:t})$ in the model of Section~\ref{section:framework}). For the rest of this paper, we assume the model to be the geological model defined in this section.  

\subsection{Distance based likelihood function}
\label{sec:likelihood}

In a complex simulator, the likelihood is intractable and an ABC filtering~\citep{jasra2012filtering} or ABC-MCMC~\citep{marjoram2003markov} methods may be useful. However, in the geological model (without the distance based likelihood) $p_{emit}(o_t|s_{t})$ and $p_{sim}(s_t|s_{t-1}, u_{t})$ are delta distributions with a single atom. Thus, the likelihood is tractable. However, the deterministic structure of the simulator can result in poor performance of any inference method. For instance, in an SMC scheme, $p_{emit}(o_t|s_{t})p_{sim}(s_t|s_{t-1}, u_{t})$, which appears in the weight updating equations of particles, will be zero for all the particles with probability one. This explains the reason behind defining the likelihood in terms of the distance between the real data and the generated observation.

Recall that, for lobe $t$ and well $\ell$, we denote the generated data by $o_{t,\ell}$ and the real data by $r_{t,\ell}$. We assume that the error values for different locations are independent. For each location we use the following kernel to compute the error $k_{\gamma}(o_{t,\ell}, r_{t,\ell})=exp(-\gamma\|o_{t,\ell} - r_{t,\ell}\|)$ where $\gamma$ is the parameter of the kernel which controls the bandwidth.

\subsection{Probabilistic programming and automatic inference}
\label{sec:venture}

We use Venture~\citep{2014arXiv1404.0099M} to represent our probabilistic model, including an interface to external simulation software. We use the automatic inference mechanisms in Venture to implement several strategies that can be applied to any simulator that satisfies the requirements of our interface. The Venture program we use for inverting all such simulators requires under 20 lines of probabilistic code.

Figure~\ref{fig:venturecode} shows the probabilistic code for the model and four inference strategies. We focus on inversion strategies built out of the building blocks provided by Metropolis Hastings (MH) and particle Markov chain Monte Carlo methods (PMCMC)~\citep{andrieu2010particle}.  
Each of the four strategies we present requires 4 or fewer lines of probabilistic code to implement. See \citep{2014arXiv1404.0099M} for details regarding the syntax and semantics of Venture. 

\begin{figure}
\vspace{-12mm}
\small\begin{Verbatim}[gobble=-30, numbers=left,numbersep=2pt, frame=single, xleftmargin=10mm,
xrightmargin = 0mm, fontsize=\relsize{-2}]
[assume sim (make_simulator) ]
[assume get_init (lambda () (sim 'initialize)) ]
[assume get_params (mem (lambda (t) (scope_include 0 t (make_array (uniform_continuous 0 1) 5)))) ]
[assume get_state (mem (lambda (t) (sim 'simulate (get_params t) t (if (= t 0)
                                                                       (get_init)
                                                                       (get_state (- t 1))))))]
[assume get_emission (mem (lambda (t) (sim 'emit (get_state t)))))]
[assume get_distance (mem (lambda (t) (sim 'distance (get_emission t)))))]
\end{Verbatim} 

\vspace{-10mm}
\begin{multicols}{2}
\small\begin{Verbatim}[gobble=-30, numbers=left,numbersep=2pt, frame=single, 
xleftmargin=10mm,
xrightmargin = -5mm, fontsize=, fontsize=\relsize{-2}]
// Particle Gibbs
for t = 1...T:
    [observe (log_flip (get_distance t)) true]
[infer (pgibbs 0 ordered 10 50)]
\end{Verbatim}
\columnbreak
\small\begin{Verbatim}[gobble=-30, numbers=left,numbersep=2pt, frame=single, 
xleftmargin=8mm,
xrightmargin = 0mm, fontsize=, fontsize=\relsize{-2}]
// Metropolis Hastings
for t = 1...T:
    [observe (log_flip (get_distance t)) true]
[infer (mh default one 500)]
\end{Verbatim} 
\end{multicols}

\vspace{-10mm}
\begin{multicols}{2}
\small\begin{Verbatim}[gobble=-30, numbers=left,numbersep=2pt, frame=single, 
xleftmargin=10mm,
xrightmargin = -5mm, fontsize=, fontsize=\relsize{-2}]
// Hybrid PGibbs-MH
for t = 1...T:
    [observe (log_flip (get_distance t)) true]
[infer (cycle ((pgibbs 0 ordered 10 10)
               (mh default one 50)) 10)]
\end{Verbatim}
\columnbreak

\small\begin{Verbatim}[gobble=-30, numbers=left,numbersep=2pt, frame=single, 
xleftmargin=8mm,
xrightmargin = 0mm, fontsize=, fontsize=\relsize{-2}]
// Sequential MH
for t = 1...T:
    [observe (log_flip (get_distance t)) true]
    for i == 1...t:
        [infer (mh 0 one 10)]
\end{Verbatim}
\end{multicols}

\vspace{-2mm}

\vspace{-4mm}
\caption{{\footnotesize{\textbf{A probabilistic program implementing our framework for inverting sequential simulators.} \emph{(Top code block)} Venture code for the probabilistic model, using a single procedure {\tt sim} to interface with external simulation software (in our experiments , via a Python to MATLAB link). \emph{(Middle left)} The code for loading in observations (e.g. from well logs) and for running a particle Gibbs method for 50 iterations and 10 particles. \emph{(Middle right)} Running single-site (random scan) MH for 500 iterations. \emph{(Bottom right)} Sequential MH with $10t$ iterations over the first $t$ observations. In this strategy, data incorporation is interleaved with inference, to incrementally account for strong sequential dependencies. \emph{(Bottom left)} A hybrid method based on alternating particle Gibbs and single-site (random scan) MH.
}
}}
\label{fig:venturecode}
%\vspace{-60mm}
\end{figure}

\section{Experiments}

\begin{figure}
\vspace{-3mm}
\begin{tabular}{cccc}
\includegraphics[trim = -0mm 0mm 11mm 0mm, clip, scale = 0.2]{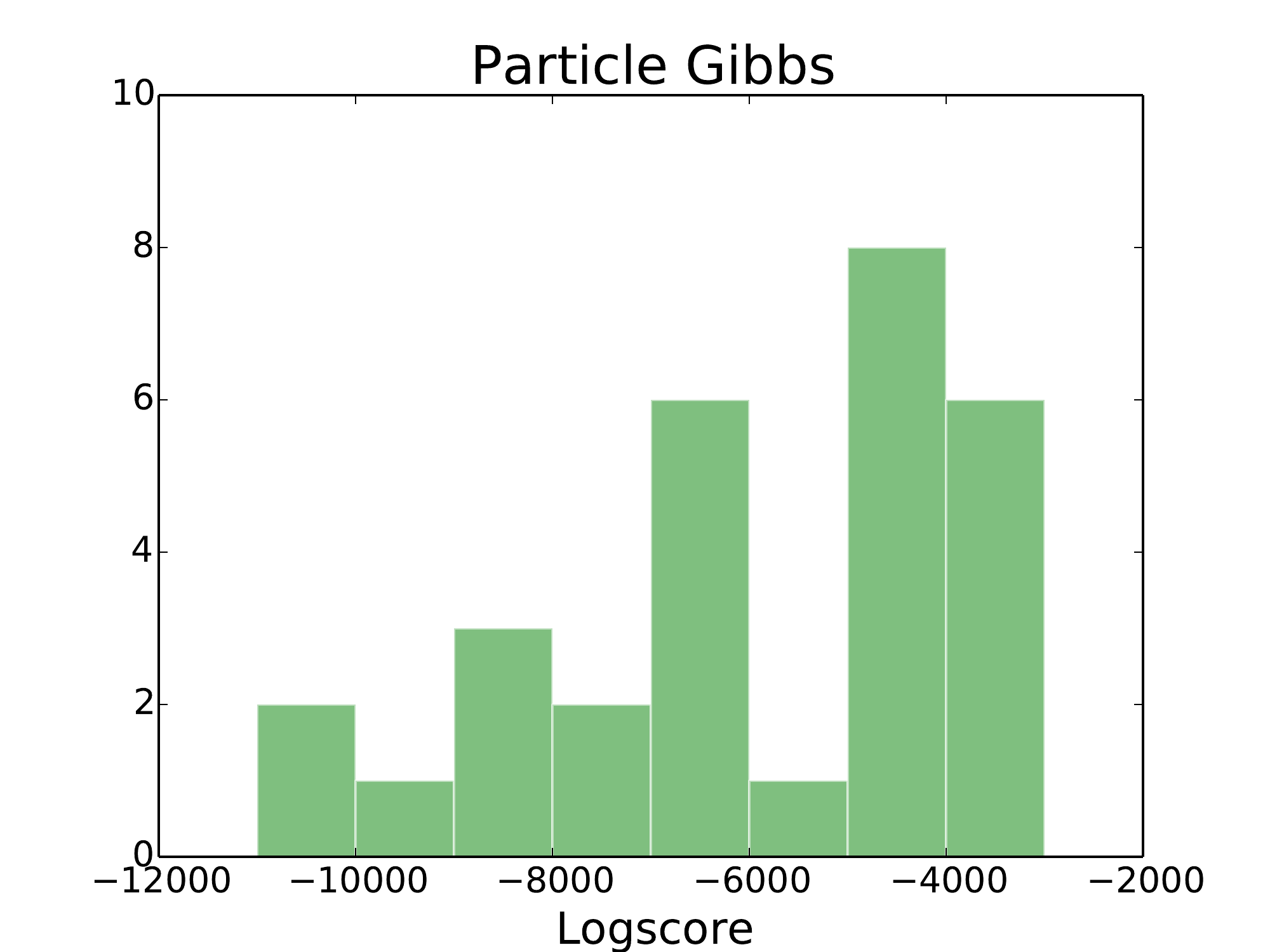} &
\includegraphics[trim = 0mm 0mm 11mm 0mm, clip, scale = 0.2]{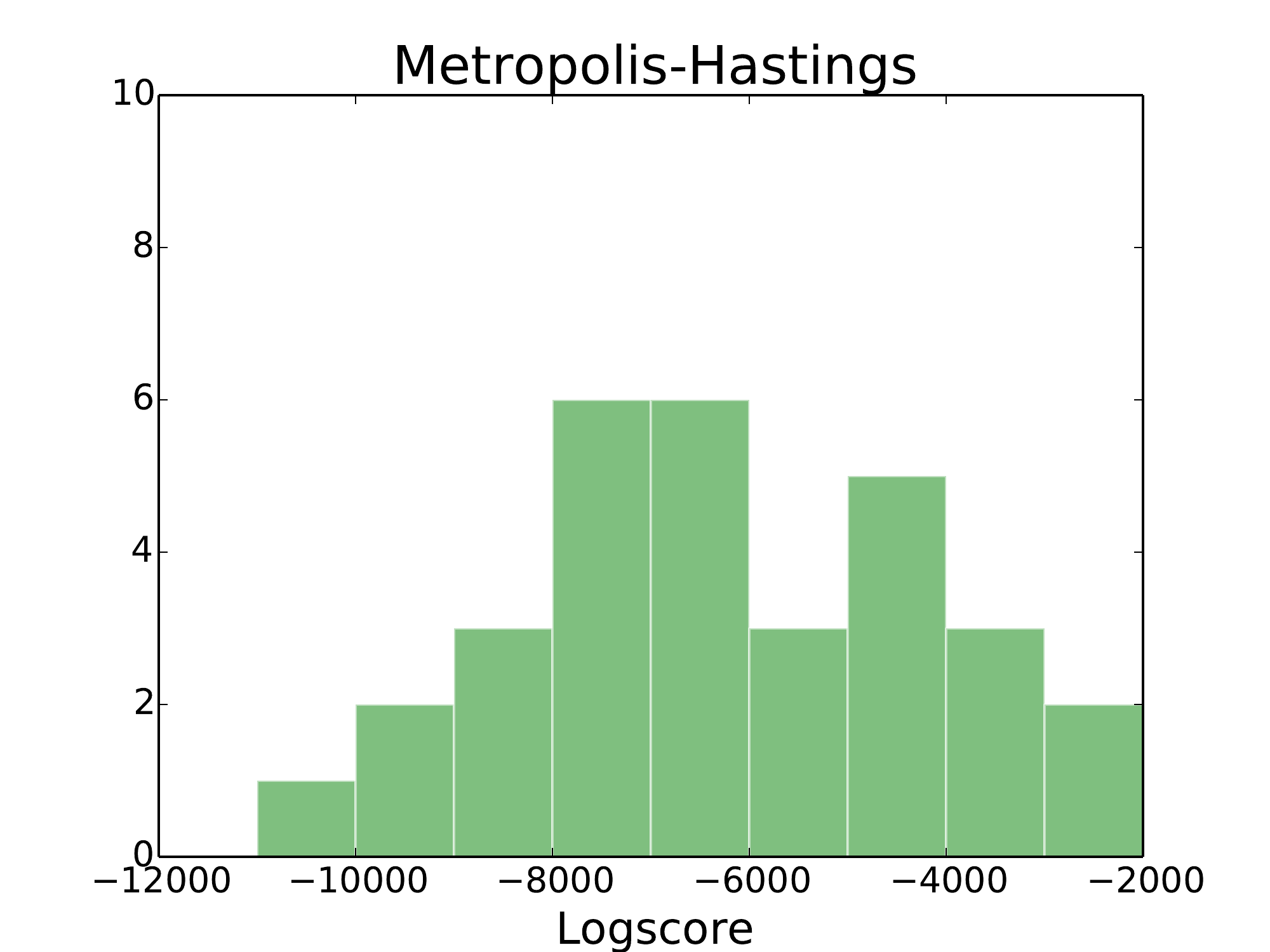}&
\includegraphics[trim = -0mm 0mm 11mm 0mm, clip, scale = 0.2]{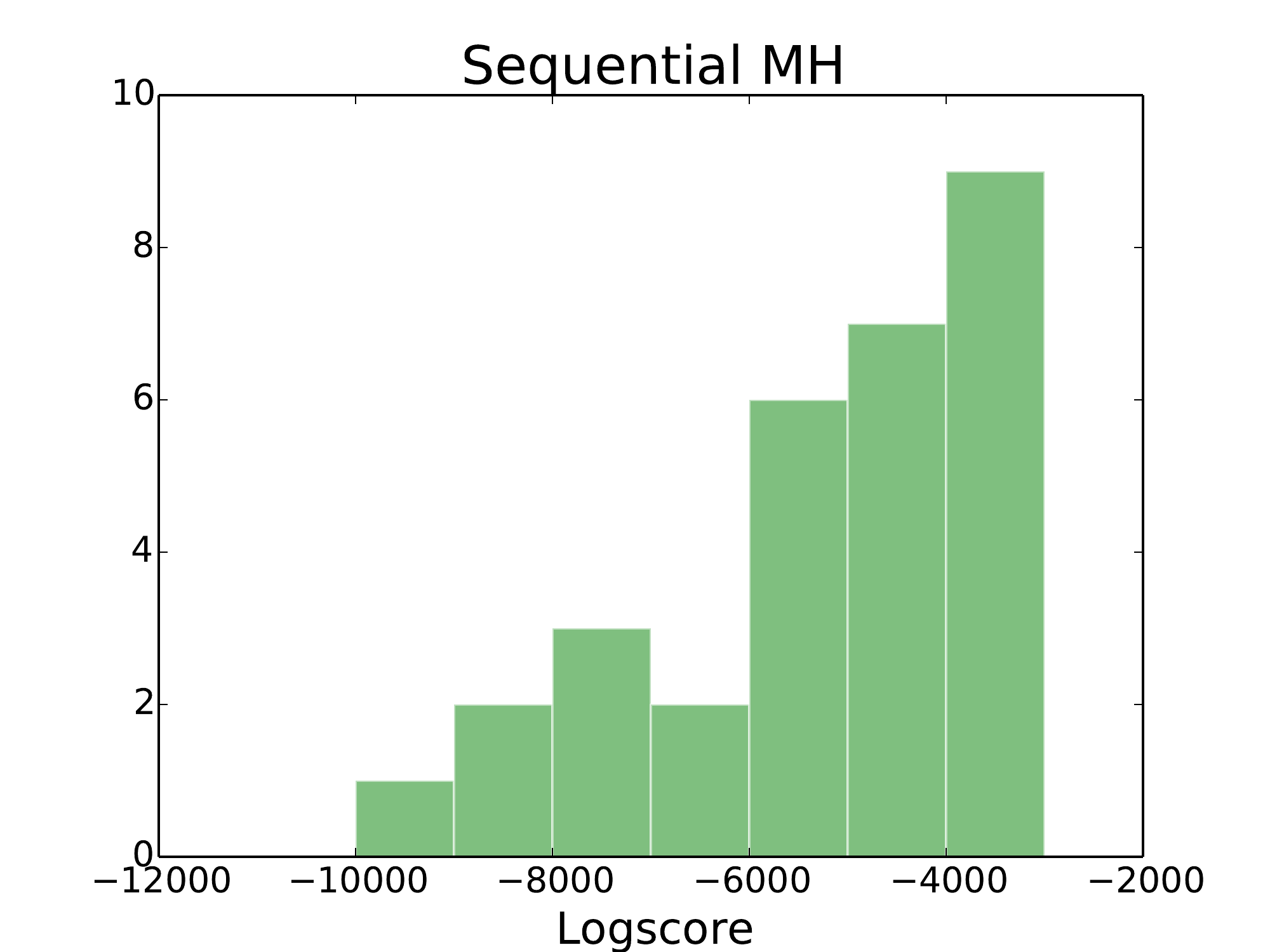}&
\includegraphics[trim = -0mm 0mm 0mm 0mm, clip, scale = 0.2]{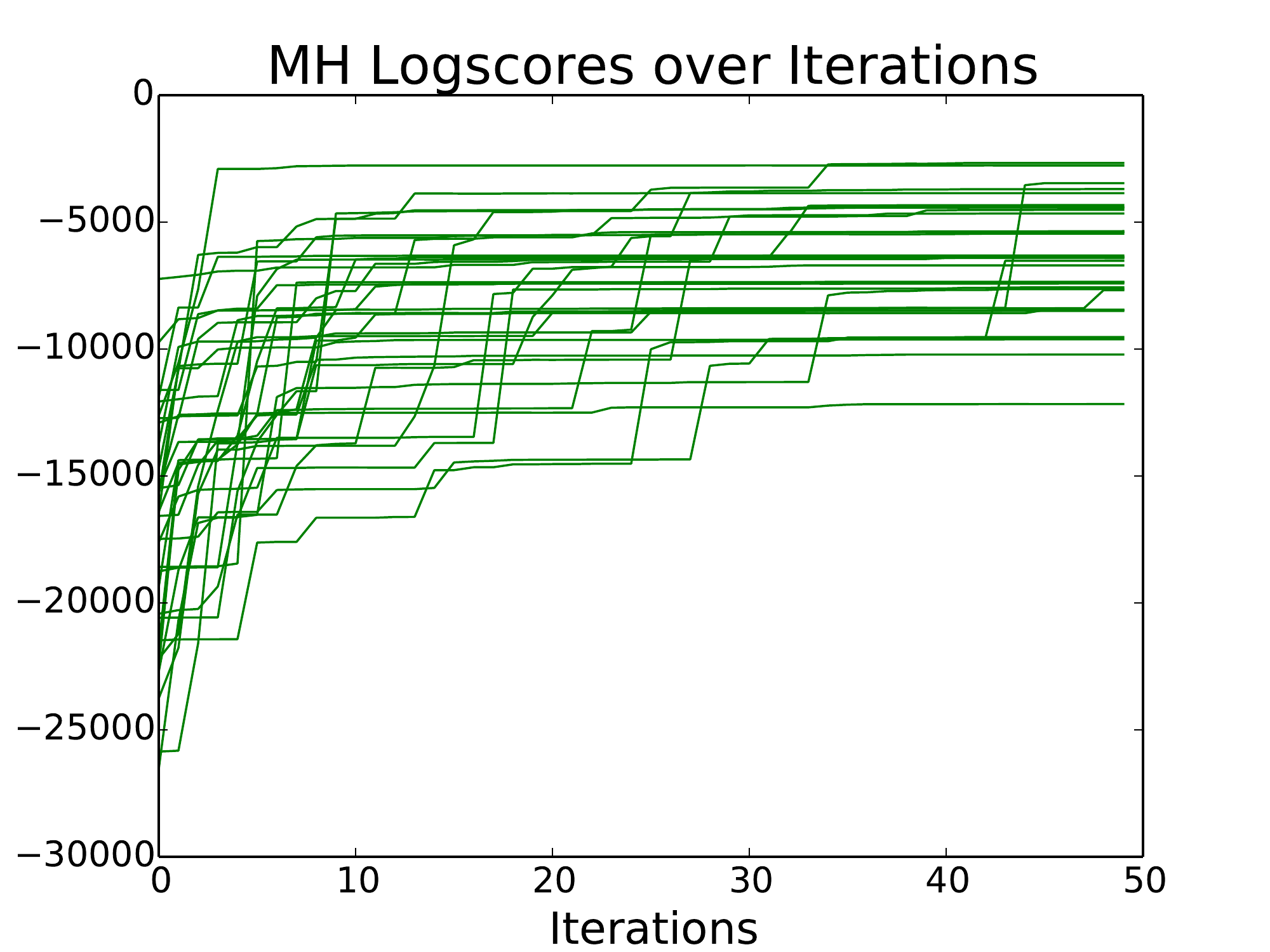}\\
\end{tabular}
\vspace{-3mm}
\caption{\footnotesize{\textbf{Comparison of automatic inference methods on the geological simulator.} We show histograms of log probability for 30 independent runs of each method. We also show the trajectories taken by the Metropolis-Hastings method.}}

\label{fig:hist}
\vspace{-6mm}
\end{figure}

We report two experimental results. First, we compare the performance of particle Gibbs, Metropolis-Hastings and sequential Metropolis-Hastings. For all three methods, we set the kernel bandwidth $\gamma$ to 1. We use 10 lobes; each of these problem instances involves exploring a jagged 50-dimensional energy landscape. Figure~\ref{fig:hist} shows the results. On this problem, we find sequential Metropolis-Hastings to be the most effective, with particle Gibbs also exhibiting reasonable performance. Pure Metropolis-Hastings occasionally performs quite well but exhibits higher variance, frequently getting stuck in local minima.

The three methods were run for roughly equivalent numbers of iterations (500), and their runtimes were all within a factor of 2 of each other. This is due to the runtime being dominated by calls to the simulator, and one iteration resulting in about 5-10 simulator calls for all three methods.

\begin{figure*}[!htb]

\begin{tabular}{ccc}
\includegraphics[trim = -6mm 10mm 100mm 50mm, clip, scale = \welllogscale]{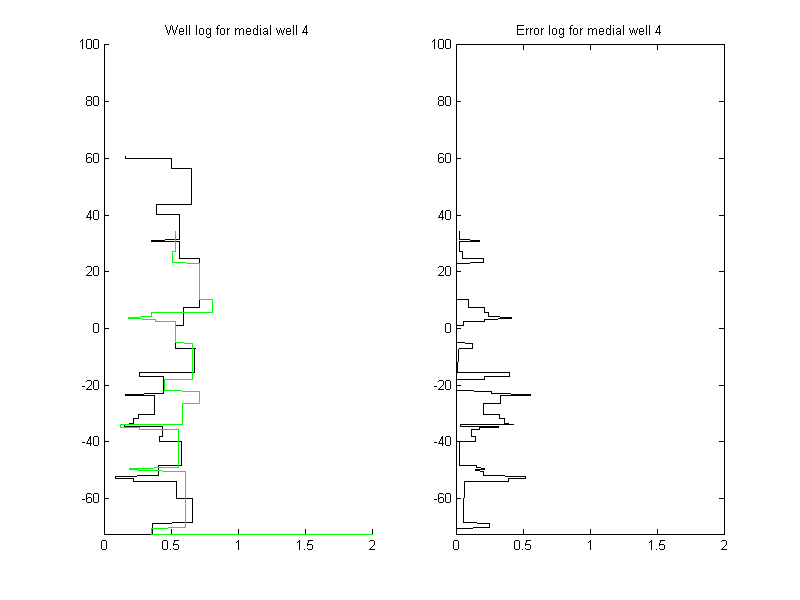} &
\includegraphics[trim = -6mm 10mm 100mm 50mm, clip, scale = \welllogscale]{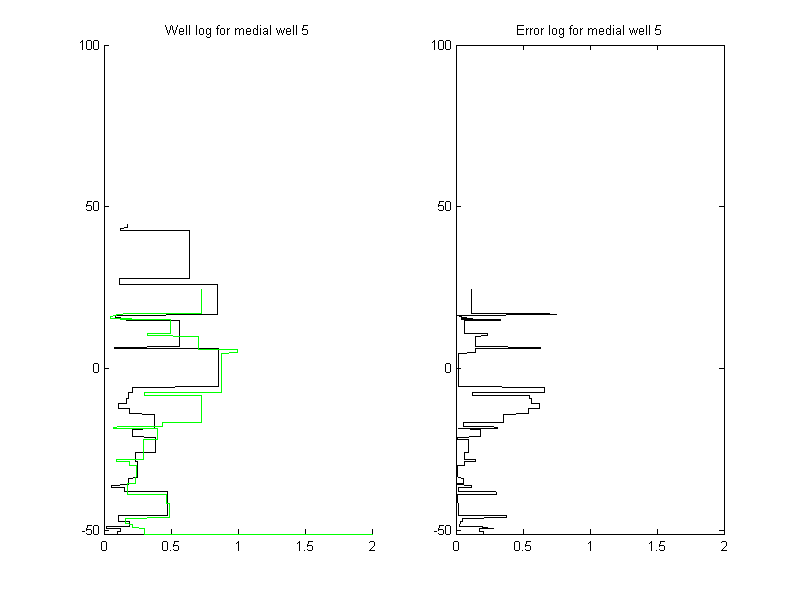} &
\includegraphics[trim = -6mm 10mm 100mm 50mm, clip, scale = \welllogscale]{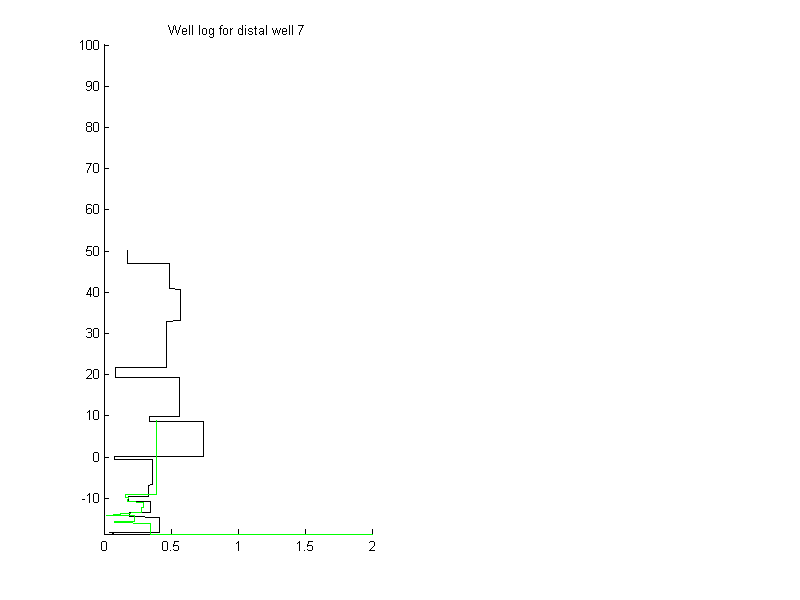}\\
\footnotesize{(a) PGibbs - Well \#4} & \footnotesize{(b) PGibbs - Well \#5} & \footnotesize{(c) PGibbs - Well \#6 } \end{tabular}

\vspace{-3mm}
\begin{comment}
\begin{tabular}{ccc}
\includegraphics[trim = 20mm 10mm 100mm 40mm, clip, scale = \welllogscale]{Figures/mh_well4_5000iter_80lob.png} &
\includegraphics[trim = 20mm 10mm 100mm 40mm, clip, scale = \welllogscale]{Figures/mh_well5_5000iter_80lob.png} &
\includegraphics[trim = 20mm 10mm 100mm 40mm, clip, scale = \welllogscale]{Figures/mh_well6_5000iter_80lob.png}\\
(f) MH - Well \#4 & (g) MH - Well \#5 & (h) MH - Well \#6 \\
\end{tabular}
\end{comment}

\caption{\footnotesize{\textbf{Typical large-scale (80 lobe) well log fits}. \emph{(a,b,c)} Inverted (green) and true (black) well logs obtained using particle Gibbs with 250 particles and 2 transitions; results correspond to the particle with highest weight.}}
\label{figA:well7_tab}
\vspace{-6mm}
\end{figure*}

Figure~\ref{figA:well7_tab} shows typical results for a larger-scale experiment searching over 80 lobes (400 dimensions). Many complex features of the well logs are captured by our method, although some wells are only poorly explained. Based on discussions with geologists, these results are comparable in quality to those obtained via a custom optimization-based baseline. It thus may be possible to improve accuracy as well as reduce code complexity by developing a more sophisticated inference scheme that can be automatically applied to this broad class of inversion problems.

\begin{comment}
\begin{figure}[h]
\vspace{-1mm}
 \centering
\begin{tabular}{c}
\includegraphics[trim = 0mm 0mm 0mm 0mm, clip, scale = 0.3]{Figures/mh_logscore_series.pdf}\\
\end{tabular}
\vspace{-3mm}
\caption{\footnotesize{Logscores for the 30 runs of MH.}  }
    \label{fig:mh_iters}
\vspace{-7mm}
\end{figure}
\end{comment}

\vspace{-1mm}

\section{Discussion}

These preliminary results show that it is possible to use a general-purpose probabilistic programming system with only automatic, general-purpose inference mechanisms to invert sophisticated software simulators. A 10-line probabilistic program suffices. Multiple inference strategies can be specified with 4 or fewer lines, and can be compared to produce an ensemble of probable inversions. If the underlying simulator is changed, the only change to the probabilistic program that is needed is to generate the appropriate random parameters per simulation step. The inference programs do not need to be changed at all, even though the transition operators they induce may be quite different.

\vspace{-3mm}
Future work will investigate more sophisticated models and automatic, general-purpose inference schemes, as well as applications to other simulators. It would be especially interesting to address statistical issues in inversion, for example by augmenting our simulator interface to expose and label parameters that affect model complexity or adjust the resolution of the data, and using model selection and parameter estimation to adjust them appropriately.

\vspace{-2mm}
\subsubsection*{Acknowledgements} 
\vspace{-2mm}
We thank Shell company for the funding and provision of the lobes forward stratigraphy simulator. We also thank Zoltan Sylvester, Alessandro Cantelli, Matt Wolinsky and Oriol Falivene who built the simulator code.
\bibliography{Venture_code}

\begin{thebibliography}{11}
\providecommand{\natexlab}[1]{#1}
\providecommand{\url}[1]{\texttt{#1}}
\expandafter\ifx\csname urlstyle\endcsname\relax
  \providecommand{\doi}[1]{doi: #1}\else
  \providecommand{\doi}{doi: \begingroup \urlstyle{rm}\Url}\fi

\bibitem[Andrieu et~al.(2003)Andrieu, De~Freitas, Doucet, and
  Jordan]{andrieu2003introduction}
Christophe Andrieu, Nando De~Freitas, Arnaud Doucet, and Michael~I Jordan.
\newblock An introduction to mcmc for machine learning.
\newblock \emph{Machine learning}, 50\penalty0 (1-2):\penalty0 5--43, 2003.

\bibitem[Andrieu et~al.(2010)Andrieu, Doucet, and
  Holenstein]{andrieu2010particle}
Christophe Andrieu, Arnaud Doucet, and Roman Holenstein.
\newblock Particle markov chain monte carlo methods.
\newblock \emph{Journal of the Royal Statistical Society: Series B (Statistical
  Methodology)}, 72\penalty0 (3):\penalty0 269--342, 2010.

\bibitem[Boschetti et~al.(1996)Boschetti, Dentith, and
  List]{boschetti1996inversion}
Fabio Boschetti, Mike~C Dentith, and Ron~D List.
\newblock Inversion of seismic refraction data using genetic algorithms.
\newblock \emph{Geophysics}, 61\penalty0 (6):\penalty0 1715--1727, 1996.

\bibitem[Calvet and Fisher(2007)]{calvet2007multifrequency}
Laurent~E Calvet and Adlai~J Fisher.
\newblock Multifrequency news and stock returns.
\newblock \emph{Journal of Financial Economics}, 86\penalty0 (1):\penalty0
  178--212, 2007.

\bibitem[Chen et~al.(2006)Chen, Hubbard, Peterson, Williams, Fienen, Jardine,
  and Watson]{chen2006development}
J~Chen, Susan Hubbard, J~Peterson, K~Williams, M~Fienen, P~Jardine, and
  D~Watson.
\newblock Development of a joint hydrogeophysical inversion approach and
  application to a contaminated fractured aquifer.
\newblock \emph{Water Resources Research}, 42\penalty0 (6), 2006.

\bibitem[Jasra et~al.(2012)Jasra, Singh, Martin, and McCoy]{jasra2012filtering}
Ajay Jasra, Sumeetpal~S Singh, James~S Martin, and Emma McCoy.
\newblock Filtering via approximate bayesian computation.
\newblock \emph{Statistics and Computing}, 22\penalty0 (6):\penalty0
  1223--1237, 2012.

\bibitem[Malinverno(2002)]{malinverno2002parsimonious}
Alberto Malinverno.
\newblock Parsimonious bayesian markov chain monte carlo inversion in a
  nonlinear geophysical problem.
\newblock \emph{Geophysical Journal International}, 151\penalty0 (3):\penalty0
  675--688, 2002.

\bibitem[{Mansinghka} et~al.(2014){Mansinghka}, {Selsam}, and
  {Perov}]{2014arXiv1404.0099M}
Vikash {Mansinghka}, Daniel {Selsam}, and Yura {Perov}.
\newblock {Venture: a higher-order probabilistic programming platform with
  programmable inference}.
\newblock \emph{ArXiv e-prints}, March 2014.

\bibitem[Marjoram et~al.(2003)Marjoram, Molitor, Plagnol, and
  Tavar{\'e}]{marjoram2003markov}
Paul Marjoram, John Molitor, Vincent Plagnol, and Simon Tavar{\'e}.
\newblock Markov chain monte carlo without likelihoods.
\newblock \emph{Proceedings of the National Academy of Sciences}, 100\penalty0
  (26):\penalty0 15324--15328, 2003.

\bibitem[Ramillien(2001)]{ramillien2001genetic}
Guillaume Ramillien.
\newblock Genetic algorithms for geophysical parameter inversion from altimeter
  data.
\newblock \emph{Geophysical Journal International}, 147\penalty0 (2):\penalty0
  393--402, 2001.

\bibitem[Symes et~al.(2011)Symes, Sun, and Enriquez]{symes2011modelling}
William~W Symes, Dong Sun, and Marco Enriquez.
\newblock From modelling to inversion: designing a well-adapted simulator.
\newblock \emph{Geophysical Prospecting}, 59\penalty0 (5):\penalty0 814--833,
  2011.

\end{thebibliography}

\end{document}